\documentclass{article}

\usepackage{microtype}
\usepackage{graphicx}
\usepackage{subfigure}
\usepackage{booktabs} 
\usepackage{amsmath}

\usepackage{hyperref}

\usepackage[accepted]{mlsys2023}
\mlsystitlerunning{Accelerating Parallel Stochastic Gradient Descent via Non-blocking Mini-batches}

\begin{document}

\twocolumn[
\mlsystitle{Accelerating Parallel Stochastic Gradient Descent via Non-blocking Mini-batches}



\mlsyssetsymbol{equal}{*}

\begin{mlsysauthorlist}
\mlsysauthor{Haoze He}{to}
\mlsysauthor{Parijat Dube}{goo}
\end{mlsysauthorlist}

\mlsysaffiliation{to}{Department of Electrical and Computer Engineering, New York University, New York, USA}
\mlsysaffiliation{goo}{IBM Research, New York, USA}

\mlsyscorrespondingauthor{Haoze He}{hh2537@nyu.edu}

\mlsyskeywords{Machine Learning, MLSys}

\vskip 0.3in

\begin{abstract}
SOTA decentralized SGD algorithms can overcome the bandwidth bottleneck at the parameter server by using communication collectives like Ring All-Reduce for synchronization. 
While the parameter updates in distributed SGD may happen asynchronously there is still a synchronization barrier to make sure that the local training epoch at every learner is complete before the learners can advance to the next epoch. The delays in waiting for the slowest learners(stragglers) remain to be a problem in the synchronization steps of these state-of-the-art decentralized frameworks. In this paper, we propose the (de)centralized Non-blocking SGD (Non-blocking SGD) which can address the straggler problem in a heterogeneous environment. The main idea of Non-blocking SGD is to split the original batch into mini-batches, then accumulate the gradients and update the model based on finished mini-batches. The Non-blocking idea can be implemented using decentralized algorithms including Ring All-reduce, D-PSGD, and MATCHA to solve the straggler problem. Moreover, using gradient accumulation to update the model also guarantees convergence and avoids gradient staleness. Run-time analysis with random straggler delays and computational efficiency/throughput of devices is also presented to show the advantage of Non-blocking SGD. Experiments on a suite of datasets and deep learning networks validate the theoretical analyses and demonstrate that Non-blocking SGD speeds up the training and fastens the convergence. Compared with the state-of-the-art decentralized asynchronous algorithms like D-PSGD and MACHA, Non-blocking SGD takes up to 2x fewer time to reach the same training loss in a heterogeneous environment.
\end{abstract}
]
\printAffiliationsAndNotice{}  
\section{Introduction}
\label{submission}

Stochastic gradient descent (SGD) is the backbone of most state-of-the-art machine learning algorithms. Hence, guaranteeing the convergence rate and stability of the SGD algorithm as well as speeding-up SGD are critical to machine learning training. Classical SGD was designed to be run serially at a single node; the stability and convergence rate of single node SGD are reliable and are well studied~\citep{dekel2012optimal, ghadimi2013stochastic}. However, the size of deep learning models, the magnitude of computations,  and the size of datasets have increased dramatically in recent years. Running SGD at a single node can be prohibitively slow. To solve this problem, distributed implementations of SGD, which parallelize the training across multiple worker nodes are shown to be successful. Following are the major types of distributed SGD: \\
\textbf{Synchronous/Asynchronous SGD with Centralized Aggregation}
The most popular framework to implement distributed SGD and parallelize gradient computation is using a centralized aggregator (aka parameter server)~\citep{dean2013tail, li2014communication, cui2014exploiting, dutta2016short}. In distributed SGD the parameter updates can be designed as either synchronous or asynchronous. During synchronization in data-parallel distributed SGD, the parameter server waits for all the nodes to finish their local training and push gradients before it updates the model parameters. In asynchronous distributed SGD, the model parameters are updated asynchronously at the parameter server when any learner pushes the gradient, thereby preventing the straggler problem.
However, synchronous distributed SGD faces the challenge of straggling learners: random delays in computation and heterogeneous environments are common in distributed systems and edge devices. Waiting for the straggling slower learners can diminish the speed-up offered by parallelizing the training. Asynchronous distributed SGD is associated with gradient staleness problem: learners may return stale gradients that were evaluated at an older version of the model, which will lead to slower converge and worse final model. In addition, both synchronous and asynchronous centralized distributed SGD suffer from the communication bottleneck problem when the framework has a large number of learner nodes. \\
\textbf{Decentralized Synchronous SGD}
To address the issue of communication bottleneck in centralized (a)synchronous SGD, an alternate is to perform decentralized aggregation without a parameter server. Both synchronous and asynchronous methods are switched to the decentralized algorithm. The most popular decentralized synchronous distributed SGD framework is \textbf{Ring All-reduce}, which is implemented in NVIDIA Collective Communications Library (NCCL) and supported by TensorFlow and PyTorch. Although ring all-reduce address the communication bottleneck issue, the heterogeneous environment of nodes remains to be the problem. \\
\textbf{Decentralized Asynchronous SGD}
In recent years, many papers proposed decentralized asynchronous SGD with novel topology structures. Most of them utilize parameter averaging instead of gradient updates, such as decentralized parallel SGD (D-PSGD)~\citep{lian2018asynchronous}. Each node only needs to average with its neighbors' model, thus reducing the communication complexity~\citep{blot2016gossip, jin2016scale, lian2017can}. Previous works in distributed optimization have extensively studied the error convergence of decentralized SGD in terms of iterations or communication rounds~\citep{duchi2011dual, jakovetic2018convergence, nedic2009distributed,scaman2018optimal,towfic2016excess,yuan2016convergence, zeng2018nonconvex}. But all of the previous works only focus on the number of iterations required to achieve a target error. Densely-connected networks, when used appropriately, give faster error convergence. However, they incur a higher communication delay per iteration, which typically increases with the maximal node degree and lead to worse wall-clock time. Other novel asynchronous decentralized algorithms with sparse-connected network topology have slower or even uncertain convergence.

\subsection{Limitation of Synchronous SGD: Heterogeneity in Multi-GPU Architectures}
Both centralized and decentralized synchronous distributed SGD suffer from poor performance in heterogeneous environments. There are two reasons for heterogeneity in multi-GPU training: performance gap among GPUs and sparse data assigned to different GPUs. Both TensorFlow and Omnivore support deep learning on heterogeneous CPU+GPU architecture~\citep{abadi2016tensorflow, hadjis2016omnivore}. \textbf{Performance gap} is significant if same work is assigned to different generations of GPUs/CPUs. Even within GPUs with the same architecture, the gap still exists. The clock rate and memory latency display oscillations on GPUs with the same model from the same vendor. The gap is amplified when multiple GPUs are integrated on the same server. Given the same training batch on a server with 4 NVIDIA V100 GPUs, the maximum gap of execution time between GPUs is as large as 32\% for a single epoch~\citep{ma2020heterogeneous}. \textbf{Sparse data}, which refers to the number of non-zeros data various among batches. Since sparse linear algebra operations are sensitive to the number of non-zero data, the execution time is different in processing across batches. Previous works propose algorithms for two different batch sizes concurrently to use in a heterogeneous CPU+GPU architecture to maximize utilization and reduce staleness for both resources~\citep{ma2020heterogeneous, masters2018revisiting}. However, these algorithms can not handle sparse data issues and performance gaps between GPUs with the same model. Manually assigning batch size is also inconvenient.

\subsection{Main Contribution}
In this paper, we propose the (de)centralized Non-blocking Synchronous parallel SGD (Non-blocking SGD) which can address stragglers caused by heterogeneous environments. In (de)centralized Non-blocking SGD framework, the original batch will be split into mini-batches locally at the learners and the model is updated based on only completed mini-batches (for which gradient is available). The decentralized Non-block SGD with a sparse connection topology can solve all the challenges listed above: stragglers (by minimizing the delays in waiting for the slowest learner), gradient staleness (by doing synchronous updates instead of asynchronous updates), and slower convergence (by only allowing once local update). More specifically, the main contribution of this paper is as follows:
\begin{itemize}
    \item \textbf{Non-blocking to deal with Stragglers:} Most popular distributed (a)synchronous SGD algorithms suffers from stragglers. We propose a novel non-blocking idea that can solve the straggler problem and implement it on both centralized and decentralized frameworks. Decentralized Non-blocking algorithms utilize Ring All-reduce, D-PSGD, and MATCHA as baselines and can achieve faster and more reliable convergence, fewer delays, and shorter wall-clock time.
    \item \textbf{General performance analysis against state-of-the-art algorithms:} We present both complexity and efficiency analysis of the decentralized Non-block SGD and compare it with the general decentralized state-of-the-art algorithm. Expected runtime per iteration, expected runtime per epoch, and expected throughput efficiency/computational efficiency are included in the paper. We also generalize the analysis by removing several common assumptions in previous work such as bounded delays, exponential distributed service times, and independence of the staleness process. We also use convergence analysis to prove that error convergence is the same or faster than baseline algorithms.
    \item \textbf{Extendable Non-blocking idea:} We propose the general Non-blocking algorithm which is extendable to most (de)centralized state-of-the-art algorithms. Furthermore, going beyond (de)centralized SGD, the Non-blocking idea is extendable to any distributed computation or consensus algorithm that requires frequent synchronizations. For instance, in federated learning systems whose edge devices have large performance gaps. The gradient accumulation in the Non-blocking idea can also help address the limited memory issue on edge devices. 
    \item \textbf{Experimental Results on Error-versus-wallclock Time Convergence:} We evaluate the Non-blocking idea in a heterogeneous environment using a suite of deep learning tasks, including computer vision tasks on CIFAR-10/100. The empirical results corroborate the theoretical analyses to demonstrate the that Non-blocking idea takes $\sim2\times$ fewer time to reach the same training loss in heterogeneous environments.
\end{itemize}

\section{Preliminaries and Related work}
\subsection{Problem Formulation}
Suppose a network has $P$ worker nodes to implement distributed SGD. The model parameters are denoted by $x$ where $x \in R^d$. Each worker node $i$ only has access to its own local training data distributed as $D_i$. The purpose of distributed SGD is to train a model by minimize the objective function $L(x)$ using $P$ worker nodes. The problem can be defined as follows: 
\begin{equation}
\min_{x\in R^d} L(x) = \min_{x\in R^d} \frac{1}{P} \sum_{i = 1}^P E_{s\sim D_i}[l_i(x;s)]
\end{equation}
where $l(x)$ is the loss function defined by the learning model and $ E_{s\sim D_i}[l_i(x;s)]$ is the local objective function at the $i$-th worker.
\subsection{Synchronous Centralized SGD}
Synchronous centralized SGD with parameter server is parallel mini-batch SGD, where workers compute stochastic gradients of the local objectives in parallel and use the averaged gradient to update model parameters after each iteration. The update rule is written as:
\begin{equation}
x_{k+1} = x_k -\eta\left[ \frac{1}{P}\sum_{i=1}^{P}g_i(x_k;\xi_i)\right]
\end{equation}
where $x_k$ is the parameters at the $k$-th iteration, $g_i(x_k;\xi_i)$ denotes the gradient descent, $\eta$ is the learning rate, and $\xi_i$ is the randomly sampled mini-batches from the local data distribution. The convergence analysis has been presented in~\cite{dekel2012optimal,bottou2018optimization}.

\subsection{Asynchronous  Centralized SGD}
The asynchronous decentralized SGD allows learner nodes to push gradients to the parameter server once they finish training and the parameter server updates the model parameters without waiting for the slower learners. The update rule can be written as:
\begin{equation}
x_{k+1} = x_k -\eta g_i(x^{i};\xi_i)
\end{equation}
where $x_k$ is the model parameters at the parameter server after the $k$-th update, $x^i$ is the parameters at the $i$-th learner node, $\eta$ is the learning rate, and $\xi_i$ is the randomly sampled mini-batches from the local data distribution. After the parameter server finishes the update, the learner node which pushes the gradient will pull the new model from the parameter server and start a new iteration. 

\subsection{Synchronous Decentralized SGD}
Synchronous Decentralized SGD such as ring all-reduce averages parameters of all learner nodes after each iteration. Each node gathers the model parameters from other nodes and performs the update locally. The update rule can be written as:
\begin{equation}
x_{k+1}^i = x_k -\eta g_i(x_k;\xi_i)
\end{equation}
\begin{equation}
x_{k+1} = \frac{1}{P}\sum_{i=1}^P x_{k+1}^i
\end{equation}
where $x_{k+1}^i$ denotes the parameters of the $i$-th learner node after the local update and $x_{k+1}$ denotes the parameters after ring all-reduce parameter averaging.

\begin{figure*}[t!]
    \centering
    \includegraphics[width=155mm]{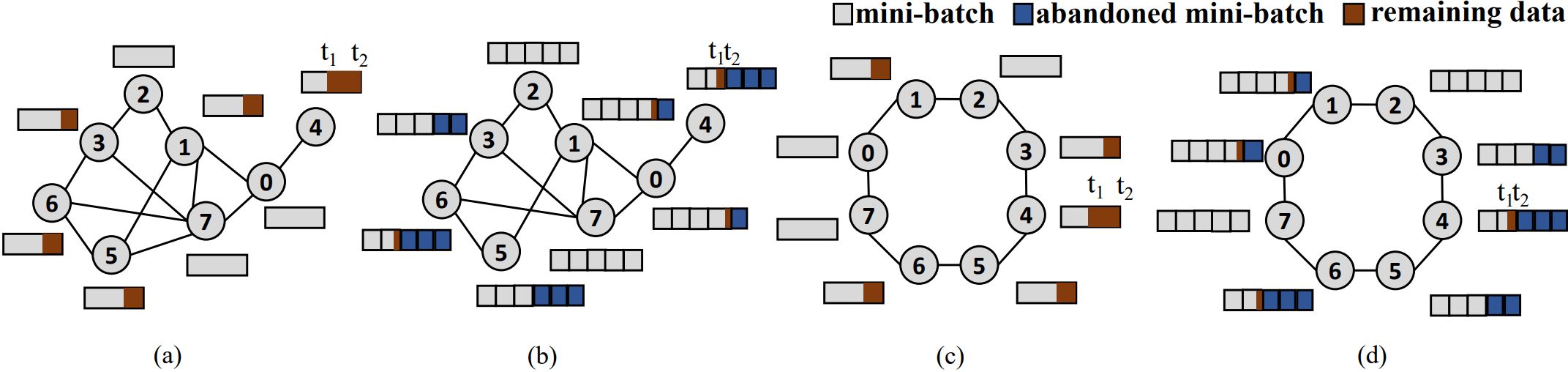}
    \caption{Illustration of Non-blocking SGD between two synchronizations. Node 7 is the fastest worker and node 4 is the slowest straggler. $t_1$ is the finish time of the fastest learner and $t_2$ is the finish time of the slowest learner. $t_2$ - $t_1$ is the time waiting for stragglers. (a) The baseline of D-PSGD algorithm using randomly generated topology. (b) The Non-blocking SGD was implemented on D-PSGD baseline. (c) The baseline of ring all-reduce (d) The Non-blocking ring all-reduce}
    \label{Fig:1}
\end{figure*}

\subsection{Asynchronous Decentralized SGD(D-PSGD)}
In recent years, many papers proposed decentralized asynchronous distributed SGD with novel topology structures.~\cite{nedic2009distributed,duchi2011dual} In D-PSGD (also referred to as consensus-based distributed SGD)~\cite{lian2017can}, nodes perform one local update and average their models only with neighboring nodes. The update rule is given as:
\begin{equation}
x_{k+1}^i = \sum_{i=1}^P W_{ij}\left[x_k^j - \eta g_j(x_k^j;\xi_j)\right]
\end{equation}
where $W \in R^{P*P}$ and $W_{ij}$ is the $(i,j)$-th element of the mixing matrix $W$ which presents the adjacency of node $i$ and $j$. $W_{ij}$ is non-zero if and only if node $i$ and node $j$ are connected. One can design a sparse mixing topology to reduce the communication complexity. Although D-PSGD has been extensively studied in the last decade and many new algorithms are proposed based on it~\cite{scaman2018optimal,wang2019matcha}, it remains open how to analyze the case when workers perform more than one local update. Since nodes only average their models with neighbors, it's slower for two non-adjacent nodes to finally converge to the same model. In addition, since D-PSGD and other state-of-the-art variants need to stop and communicate after all nodes finish certain iterations, the stragglers remain to be the problem.

\section{Proposed Method}

\subsection{General Non-blocking Algorithm}
In this section, we introduce the Non-blocking SGD algorithm. Since the stragglers' problem is caused by the performance gap between different GPUs in heterogeneous environments, allowing GPUs with different computational performance to process different amount of data before synchronization steps can reduce the proportion of waiting time and address the stragglers issue. The general non-blocking algorithm is presented in Algorithm~\ref{Alg:1}. Following Algorithm~\ref{Alg:1}, the Non-blocking idea can be applied to almost all the parallel SGD frameworks.

\begin{figure*}[t!]
    \centering
    \includegraphics[width=130mm,scale=0.5]{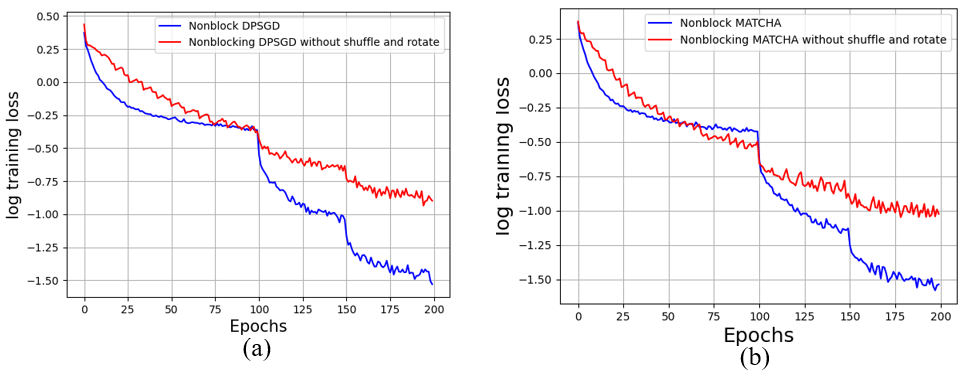}
    \caption{Illustration of training without shuffling on CIFAR-10 and CIFAR-100.}
    \label{Fig:2}
\end{figure*}

\begin{algorithm}[t!]
    \begin{algorithmic}[1]
    \caption{General Non-blocking algorithm}
    \label{Alg:1}
    \STATE \textbf{Initialization:}initialize local models $\{x_0^i\}_{i=1}^P$ with the same initialization, learning rate $\eta$, batch size $B$, and the total number of iterations $K$. Split the original dataset into P equal-sized subsets $\xi_1, \ldots, \xi_P$. Assign subset $\xi_i$ to $i$-th worker. Shuffle local subset after each epoch. \footnotemark[3] \;
    \FOR{k=0, 1, 2, ... K-1 $\leq$ K}
          \STATE Randomly sample $\xi_{k,i}$ from local subset in the $i$-th worker and divide the sample batch $\xi_{k,i}$ into N mini-batches $\xi_{k, i,0},...,\xi_{k,i,N-1}$ \;
          \FOR{n = 0, \ldots, $N-1$}
                \IF{no terminal signals detected}
                    \STATE Compute and accumulate the local stochastic gradient $\nabla F_i(x_{k, i}; \xi_{k, i,j})$ on all nodes\;
                    \STATE Broadcast terminal signals if $n=N-1$\;
                \ELSE
                    \STATE Break and abandon unfinished mini-batches \;
                \ENDIF
          \ENDFOR \footnotemark[1]
          \STATE Communication with other workers and update the local model by fetching models from neighbors\footnotemark[2]\;
            \STATE Shuffle local subset by the end of epoch\footnotemark[3]\;
    \ENDFOR
    \STATE \textbf{Output:} The average of all workers $\frac{1}{P} \sum_{i=1}^P x_{k-1,i}$\;
    \end{algorithmic}
\end{algorithm}
\begin{algorithm}[t!]
    \begin{algorithmic}[1]
    \caption{Non-blocking D-PSGD algorithm}
    \label{Alg:2}
    \STATE \textbf{Initialization:} initialize local models $\{x_0^i\}_{i=1}^P$ with the same initialization, learning rate $\eta$, batch size $B$, weight matrix $W$, and the total number of iterations $K$. Split the original dataset into P equal-sized subsets $\xi_1,\ldots, \xi_P$. Assign subset $\xi_i$ to $i$-th worker. Shuffle local subset after each epoch. \footnotemark[3]\;
    \FOR{k=0, 1, 2, ... K-1 $\leq$ K}
          \STATE Randomly sample $\xi_{k,i}$ from local subset in the $i$-th worker and divide the sample batch $\xi_{k,i}$ into N mini-batches $\xi_{k, i,1},...,\xi_{k,i,N}.$\;
          \FOR{ n = 1, \ldots, $N$}
          \IF{no terminal signals detected}
              \STATE Compute and accumulate the local stochastic gradient $\nabla F_i(x_{k, i}; \xi_{k, i,n})$ on all nodes\;
              \STATE Broadcast terminal signals if $n=N$\;
          \ELSE
              \STATE Break and abandon unfinished mini-batches \;
          \ENDIF
          \ENDFOR \footnotemark[1] 
          \STATE Compute the neighborhood weighted average by fetching optimization variables from neighbors\footnotemark[2]: $x_{k+\frac{1}{2}, i}=\sum_{p=1}^P W_{i,p}x_{k,p}$\;
          \STATE Update the local model using accumulated gradient $x_{k+1,i} \gets x_{k+\frac{1}{2}} - \eta \frac{n}{N} \sum_{j=1}^n\nabla  F_i(x_{k, i}; \xi_{k, i,j})$
          \STATE Shuffle local subset by the end of epoch\footnotemark[3]\;
    \ENDFOR 
    \STATE \textbf{Output:} The average of all workers $\frac{1}{P} \sum_{i=1}^P x_{k-1,i}$
     \end{algorithmic}
\end{algorithm}
\footnotetext[1]{Note that Line 4-7 can run asynchronous in different workers to reduce stragglers.}
\footnotetext[2]{Note that Line 8 differs when using different baseline algorithms. It can be adjusted according to specific topology structure and communication method.}
\footnotetext[3] {Shuffling step, details in section 3.2 and Fig.~\ref{Fig:2}}

\subsection{Non-blocking D-PSGD Algorithm}
\textbf{Implementation Detail:} Figure~\ref{Fig:1} briefly illustrates the Non-blocking algorithm using D-PSGD as the baseline. We can easily see that by applying Non-blocking SGD with mini-batches on different workers, the proportion of time waiting for stragglers reduces dramatically. The pseudo-code of Non-blocking D-PSGD is shown in algorithm~\ref{Alg:2}. 


\begin{algorithm}[t!]
    \begin{algorithmic}[1]
    \caption{Non-blocking Synchronous Centralized algorithm}
    \label{Alg:3}
    \STATE \textbf{Initialization:} initialize local models $\{x_0^i\}_{i=1}^P$ and the model in parameter server $x_0^{ps}$with the same initialization, learning rate $\eta$, batch size $B$, weight matrix $W$, and the total number of iterations $K$. Split the original dataset into P equal-sized subsets $\xi_1,\ldots, \xi_P$. Assign subset $\xi_i$ to $i$-th worker. Shuffle local subset after each epoch. \;
    \FOR{k=0, 1, 2, ... K-1 $\leq$ K}
          \STATE Randomly sample $\xi_{k,i}$ from local subset in the $i$-th worker and divide the sample batch $\xi_{k,i}$ into N mini-batches $\xi_{k, i,1},...,\xi_{k,i,N}.$\;
          \FOR{ n = 1, \ldots, $N$}
          \IF{no terminal signals detected}
              \STATE Compute and accumulate the local stochastic gradient $\nabla F_i(x_{k, i}; \xi_{k, i,n})$ on all nodes\;
              \STATE Broadcast terminal signals if $n=N$\;
          \ELSE
              \STATE Break and abandon unfinished mini-batches \;
          \ENDIF
          \ENDFOR 
          \STATE Update the parameter server by fetching gradient from workers: $x_{k+1, ps} = x_{k, ps} -\sum_{i=1}^P[\eta \frac{n_{k,i}}{N} \sum_{j=1}^{n_{k,i}}\nabla  F_i(x_{k, i}; \xi_{k, i,j})]$\;
          \STATE Update the local model $x_{k+1, i} \gets x_{k+1, ps}$
          \STATE Shuffle local subset by the end of epoch\;
    \ENDFOR 
    \STATE \textbf{Output:} The model in the parameter server $x_{k-1,ps}$
     \end{algorithmic}
\end{algorithm}
We briefly describe the implementation detail of the Non-blocking D-PSGD algorithm below:
\begin{itemize}
    \item \textbf{Sample data:} Split original dataset into $P$ subsets equally. Each local worker chooses a subset for training. In the same epoch, training subsets should be different among workers. Sample $\xi_{k,i}$ from local training data of the $i$-th node and divide the sample batch $\xi_{k,i}$ into $N$ mini-batches.
    \item \textbf{Adjust learning rate:} Set the initial learning rate and adjust the learning rate according to the number of finished mini-batches. The purpose of adjusting learning rate is to assign larger learning rate to learner which has processed more mini-batches in the same iteration. $n_{k,i}$ is the amount of trained mini-batches before synchronization in iteration $k$ and node $i$.  $\eta_{k,i} = \eta * \frac{n_{k,i}}{N}$. 
    \item \textbf{Compute gradients:} Compute and accumulate all the gradients of the mini-batches $\sum_{j=1}^{n_{k,i}}\nabla  F_i(x_{k, i}; \xi_{k, i,j})$. Since stragglers train fewer mini-batches under the same batch time, unfinished mini-batches will be {\it abandoned} by the stragglers.
    \item \textbf{Averaging:} Differs in different baseline algorithms. In Non-blocking D-PSGD, randomly select a neighbor worker and average the local model with the neighbor model until all the neighbors are included, i.e., $x_{k+\frac{1}{2}, i}=\sum_{p=1}^P W_{i,p}x_{k,p}$.
    \item \textbf{Gradient update:} In Non-blocking D-PSGD, the local model updated by $x_{k+1,i} \gets x_{k+\frac{1}{2}} - \eta \frac{n_{k,i}}{N} \sum_{j=1}^{n_{k,i}}\nabla  F_i(x_{k, i}; \xi_{k, i,j})$.
    \item \textbf{Shuffling:} Since stragglers will abandon unfinished mini-batches, shuffling the data guarantee all the data be well-trained. Without shuffling, the performance of the model will be worse. In Figure~\ref{Fig:2} we illustrate the importance of shuffling. The dataset cannot be fully trained without shuffling.
\end{itemize}

Note that the Non-blocking idea can also be implemented on synchronous centralized SGD. In algorithm~\ref{Alg:3} we present the pseudo-code of Non-blocking synchronous centralized D-PSGD. 

\section{Theoretical Performance Analysis}
In this section, we analyze the complexity and efficiency of the Non-blocking SGD and compare it with the majority of state-of-the-art decentralized SGD algorithms which have synchronization steps. Both D-PSGD and MATCHA are included in the comparisons. In the following subsection, different analyses will be included: in the first part, we discuss the theoretical run-time per iteration. In the second part, we discuss the expected computational efficiency/throughput efficiency. In the third part, we discuss the expected run-time per epoch. All the following run-time analyses consider exponential random straggler delays same as previous works~\cite{dutta2018slow,lee2017speeding,mitliagkas2016asynchrony, hannah2017more}.

\subsection{Definition and Notations}
\begin{itemize}
    \item $D$ denotes the dataset size
    \item $B$ denotes the batch size, $b$ denotes the mini-batch size
    \item $X_{1:P}$ denotes the first order statistic of $P$ i.i.d. random variable $X_1, X_2, ..., X_P.$
    \item $X_{K:P}$ denotes the $K^{th}$ statistic of P i.i.d. random variables $X_1, X_2, ..., X_P$
    \item $Y$ denotes a random variable and $Y= \sum_{i=1}^{P}\frac{1}{X_{i:P}}$
    \item $X_{k, i:P}$ denotes the $i^{th}$ order statistic of $P$ in $k^{th}$ iteration. i.i.d. random variable $X_1, X_2, ..., X_P$ 
    \item $\lambda$ denotes the exponential distribution parameter. We assume the wall clock time of each learner to process a single mini-batch be i.i.d. exponential random variables $X_1, ...,X_i, . . ,X_P \sim exp(\lambda) $
    \item For epoch level analysis, a (pseudo) epoch for Non-blocking SGD is the time to complete $D/(P*B)$ iterations. Note that the amount of data processed in an epoch of Non-blocking SGD may be less than $D$ as some (slow) learners are abandoning mini-batches in each iteration. 
\end{itemize}

\subsection{Theoretical Run-time Analysis}
\textbf{Lemma 1} The expected run-time per iteration for Non-block SGD is, 
\begin{equation}
E[T] = E[X_{1:P}]
\end{equation}
To proof of Lemma 1, We assume that the $P$ learners have an i.i.d. computation times. When all the learners start together, and we wait for the first learner among $P$ i.i.d. random variables to finish, the expected computation time for that iteration is $E[X_{1:P}]$. As for a total number of $J$ iterations, the expected run-time is given by \textit{J}$E[X_{1:P}]$. The run-time of fully-synchronous SGD is $E[T] = E[X_{P:P}]$. The run-time of asynchronous SGD is $E[T] = E[X_{1:P}]$, which is the same as Non-blocking synchronous SGD~\cite{dutta2018slow}.

Then, we compare the theoretical wall clock runtime of the Non-blocking decentralized SGD and D-PSGD/MATCHA to illustrate the speed-up offered by our algorithm. 
\newline\textbf{Theorem 1}.Let the wall clock time of each learner to process a single mini-batch be i.i.d. random variables $X_1,X_2, . . . ,X_P$. For the Non-blocking algorithm, the original batch will be split into $N$ mini-batches for each learner. Then the ratio of the expected run-times per iteration for Non-blocking decentralized SGD and D-PSGD/MATCHA is
\begin{equation}
\frac{E[T_{Nonblocking}]}{E[T_{DPSGD}]} = \frac{E[X_{1:P}]}{E[X_{P:P}]}
\end{equation}
\textbf{Theorem 2}.Let the wall clock time of each learner to process a single mini-batch be i.i.d. exponential random variables $X_1, ... ,X_i, . . ,X_P \sim exp(\lambda) $. $X_1$ is the time of the fastest learner. For the Non-blocking algorithm, the original batch will be split into $N$ mini-batches for each learner. The expected run-time per iteration for Non-blocking synchronous SGD and D-PSGD/MATCHA is,
\begin{equation}
E[T_{Nonblocking}] = \frac{1}{\lambda*P}
\end{equation}
\begin{equation}
E[T_{DPSGD}] = \frac{P}{\lambda} \sum_{k=0}^{P-1} \binom{P-1}{k}\frac{(-1)^k}{(k+1)^2}
\end{equation}
The full proof is provided in Appendix A.
\newline The ratio of wall-clock run-time per epoch for Non-blocking SGD and D-PSGD is the same as the ratio of the expected run times per iteration in Theorem 1. The wall-clock run-time per epoch is
\begin{equation}
E[T_{Nonblocking}] = \frac{D}{P*B}E[X_{1:P}]
\end{equation}
\begin{equation}
E[T_{DPSGD}] = \frac{D}{P*B}E[X_{P:P}]
\end{equation}
When the run-time of each worker is under exponential distribution assumption, the run-time per epoch of Non-blocking SGD and D-PSGD/MATCHA are: 
\begin{equation}
E[T_{Nonblocking}] = \frac{D}{\lambda*P^2*B}
\end{equation}
\begin{equation}
E[T_{DPSGD}] = \frac{D}{\lambda*B} \sum_{k=0}^{P-1} \binom{P-1}{k}\frac{(-1)^k}{(k+1)^2}
\end{equation}
We assume that in the last batch of an epoch, each worker will be assigned $B$ amount of data as the original batch size. 
\subsection{Expected Computational Efficiency Analysis}
\textbf{Theorem 3}.Throughput of Non-blocking SGD is
\begin{equation}
\sum_{i=1}^{P}\frac{B}{X_{i:P}}= B*Y
\end{equation}
{\it Proof:} The throughput of the distributed deep learning system is summing up the throughput of all the learners. The throughput of learner $i$ is $\frac{B}{X_{i:P}}$. 

We assume that during an iteration, each worker processes the data at a constant speed within an iteration. We then have the following: 
\newline\textbf{Corollary 1}
The total amount of data being trained in each iteration is,
\begin{equation}
\sum_{i=1}^{P}\frac{B}{X_{i:P}} * X_{1:P} = B*Y*X_{1:P}
\end{equation}
We assume the amount of data to be a fractional value.  
\newline\textbf{Theorem 4}.Throughput of D-PSGD/MATCHA is
\begin{equation}
Throughput_{DPSGD} = \frac{P*B}{X_{P:P}}
\end{equation}
\textbf{Theorem 5}. The ratio of the throughput for Non-blocking SGD and D-PSGD/MATCHA is
\begin{equation}
\frac{\sum_{i=1}^{P}\frac{B}{X_{i:P}}}{P*B} * X_{P:P} = =\frac{Y*X_{P:P}}{P}
\end{equation}
\section{Error Convergence Analysis}
To prove the Error Convergence of Non-blocking idea, two demonstrations are included in the following. In the first part, we will prove that the accumulated gradient of mini-batches is equal to the gradient of a large batch. In the second part, we will demonstrate that the average model after being updated on the accumulated gradient has no difference from the original convergence. 
\subsection{Gradient Accumulation}
The gradient accumulation and the local update based on the accumulated gradient are:
\begin{equation}
\delta x_{k,i} = -\frac{n_{k,i}}{N} \eta \sum_{j=1}^{n_{k,i}} \delta_{x_{k,i}} L(x_{k,i};\xi_{k,i,j})
\end{equation}
\begin{equation}
x_{k+1,i} = x_{k,i} + \delta x_{k,i}
\end{equation}
$\eta$ denotes the learning rate set for the original large batch, $N$ denotes the number of mini-batches,   $n_{k,i}$ denotes the number of finished mini-batches and $x_{k,i}$ denotes the local model in $k$-th iteration and $i$-th node. Randomly sample $\xi_{k,i}$ from local data $\xi_i$ and divide the sample batch $\xi_{k,i}$ into N mini-batches $\xi_{k, i,1},..., \xi_{k,i,j}, ..., \xi_{k,i,N}$.
When $n_{k,i} = N$, Equation 20 is:
\begin{equation}
\delta x_{k,i} = -\eta  \sum_{j=1}^N \delta_{x_{k,i}} L(x_{k,i};\xi_{k,i,j})
\end{equation}
The gradient on a large batch is:
\begin{equation}
\delta x_{k,i} = -\eta * \delta_{x_{k,i}} L(x_{k,i};\xi_{k,i})
\end{equation}
\begin{equation}
\begin{aligned}
\sum_{j=1}^N \delta_{x_{k,i}} L(x_{k,i};\xi_{k,i,j}) = \delta_{x_{k,i}} L(x_{k,i};\xi_{k,i})
\end{aligned}
\end{equation}
we can prove that the update based on an accumulated gradient of multiple mini-batches is equal to the update based on a gradient of a joint dataset of all finished mini-batches. There is only one local update before synchronization. This is the precondition for the convergence analysis in the next step.

\begin{figure*}[t!]
    \centering
    \includegraphics[width=130mm]{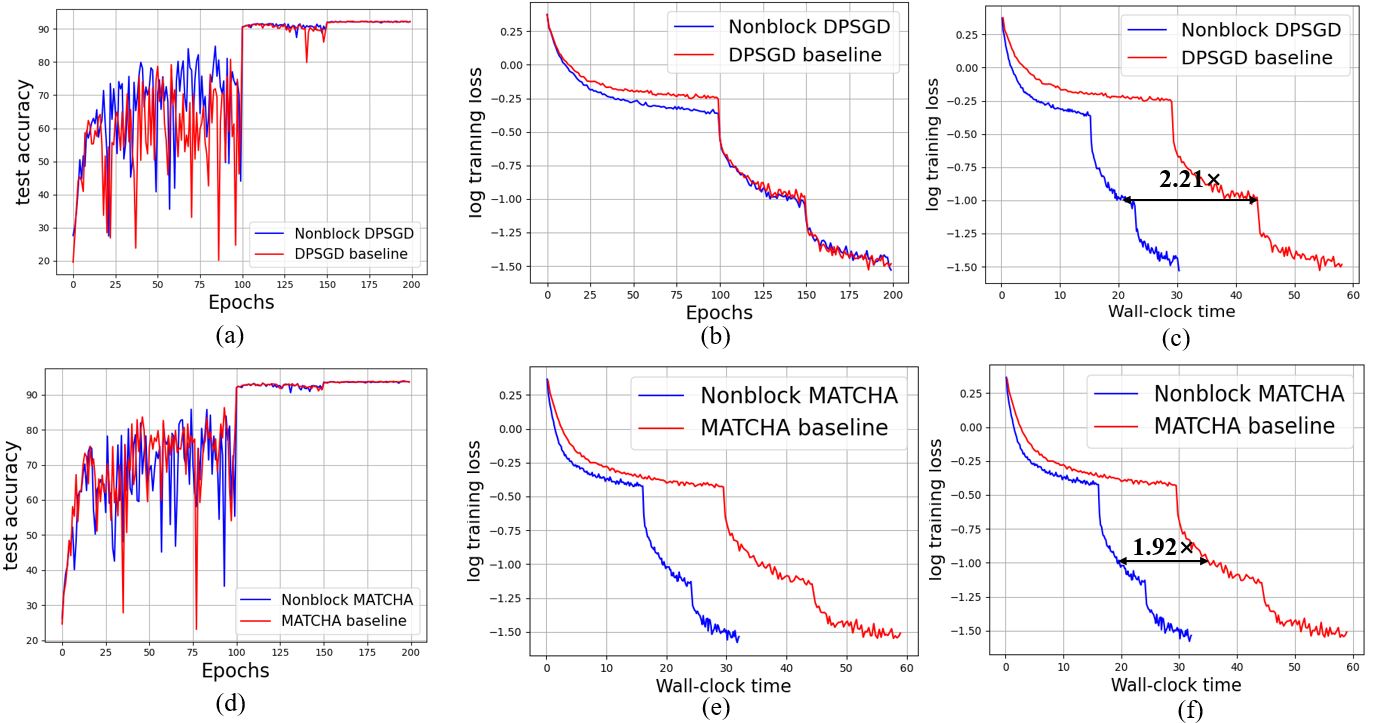}
    \caption{Performance comparison of Non-blocking algorithms and their baseline on CIFAR-10 dataset.}
    \label{Fig:3}
\end{figure*}

\begin{figure*}[t!]
    \centering
    \includegraphics[width=130mm]{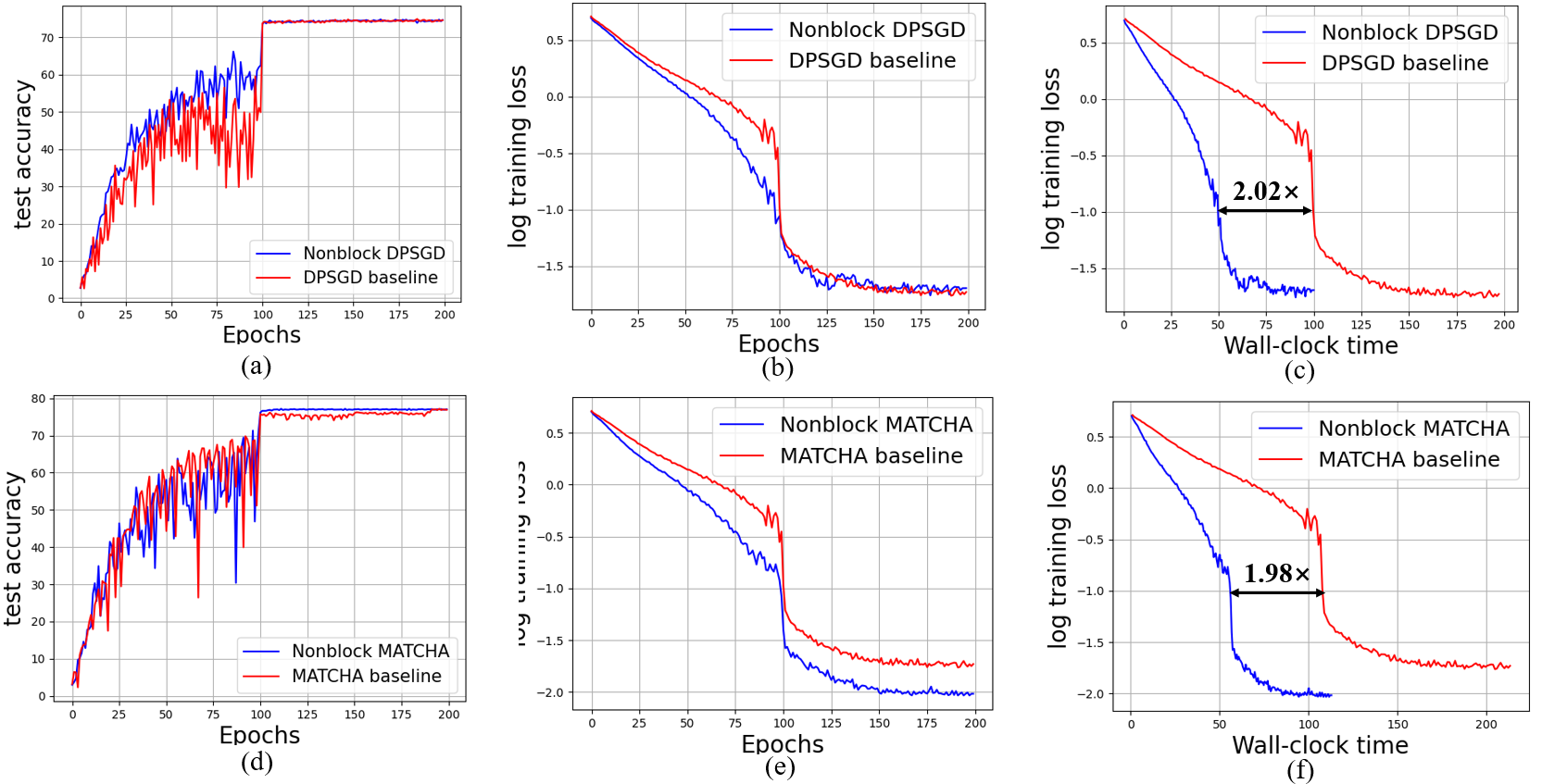}
    \caption{Performance comparison of Non-blocking algorithms and their baseline on CIFAR-100 dataset.}
    \label{Fig:4}
\end{figure*}

\subsection{Convergence rate analysis}
As for centralized synchronous distributed SGD algorithms, they gather all gradients on the parameter server(PS) and update the model on PS. Since the Non-blocking idea doesn't allow multiple local updates, the convergence of Non-blocking centralized SGD is guaranteed.  
\newline As for decentralized algorithms, we will prove the convergence analysis based on Non-blocking DPSGD. Note that this convergence analysis is a general one and can also be applied to Nonblocking MATCHA. The analysis is centered around the following assumptions:

\textbf{Assumption 1:} Each worker's local objective function $F_i(x)$is differentiable and its gradient is L-Lipschitz:$\vert\vert \nabla F_i(x) - \nabla F_i(y)\vert\vert \leq L\vert\vert x-y\vert\vert, \forall i \in \{1,2,...,P\}$

\textbf{Assumption 2:} The deviation of averaged local objectives’ gradients are bounded by a non-negative constant:$\frac{1}{P}\sum_{i=1}^P\vert\vert \nabla F_i(x) - \nabla F(x) \vert\vert^2 \leq \zeta^2$

\textbf{Assumption 3:} The variance of stochastic gradients $E\{\vert\vert g_i(x^{(k)};\xi^{(k)}) - \nabla F_i(x^{(k)})\vert\vert\}$ at any worker node is bounded for $\xi$ from the distribution $D_i$. This implies there exist constants $\sigma, \psi$ such that $\forall x_i\in \{x_1, ..., x_P\}$
$$
\begin{aligned}
&\vert\vert g(x_i^{(k)};\xi^{(k)}) - \nabla F(x_i^{(k)})\vert\vert \leq \sigma^2,  \\
&\vert\vert\nabla F(x_i^{(k)}) - \nabla F(x)\vert\vert^2 \leq \psi^2, \forall x.  \\
\end{aligned}
$$
\newline Larger batch size in local worker will lead to tighter bound. In this case, we assume $\psi$=0.

\textbf{Assumption 4:} Stochastic gradients at each worker node are unbiased estimates of the true gradient of the local objectives: $ E[F_i(x_{k, i}; \xi_{k, i})| F^{(k)}]  = \nabla  F_i(x_{k, i}; \xi_{k, i}), \forall i \in \{1,2,\ldots,P\}$.  $F^{k}$ denotes the sigma algebra generated by noise in the stochastic gradients in iteration k.

\textbf{Theorem 6: (Convergence of Non-blocking D-PSGD)}: Our convergence analysis is under Assumptions of 1 to 4. In Nonblocking DPSGD, the learning rate is adjusted according to the finished mini-batches. Assume the learning rates in different workers are $\frac{1}{2}\{\frac{1}{2L + \rho \sqrt{K/n}}\}^3  \leq \eta_1L, \eta_2L, \ldots, \eta_PL \leq  \frac{1}{2L + \rho \sqrt{K/n}} ^3$ and the fastest worker is at most one time faster than the slowest one. Then we have the following convergence rate for Nonblocking DPSGD:

$$
\begin{aligned}
&(\frac{D_1}{K} + 1-\frac{1}{(2L+\sigma\sqrt{K/P})})\sum_{k=0}^{K-1}E||\nabla f(\frac{1}{P}\sum_{i=1}^PX_{k,i})||^2  \\ 
&\leq (\frac{1}{4P+2P\sigma\sqrt{K/P}})^2 (\frac{L^2P\sigma^2}{(1-\rho)D_2} + \frac{9L^2P\varsigma^2}{(1-\sqrt{\rho})^2D_2})\\
&+ \frac{L}{8P+4P\sigma\sqrt{K/P}} + \frac{(f(0)-f^*)(2L+\sigma\sqrt{K/P})}{K}  \\
\end{aligned}
$$
Note that to simplify the formula we set
\begin{equation*}
    D_1=(\frac{1}{2}-\frac{9\eta^2L^2P}{(1-\sqrt{\rho})^2D_2}); D_2 = (1-\frac{18\eta^2}{(1-\sqrt{\rho})^2}PL^2)
\end{equation*}
When iteration $K$ sufficient large enough, we can simplify the convergence rate of Nonblocking DPSGD to:
$$
\begin{aligned}
&\frac{1}{K}\sum_{k=0}^{K-1}E||\nabla f(\frac{1}{P}\sum_{i=1}^PX_{k,i})||^2 \leq \\
&\frac{16(f(0) - f^*)L}{K} + \frac{(16f(0) - 16f^* + 8L)\sigma}{\sqrt{KP}} \\
\end{aligned} 
$$
This suggest that the convergence rate for Nonblocking D-PSGD is bounded by$O(\frac{1}{K} + \frac{1}{\sqrt{PK}})$. In addition, when iteration $K$ is large enough, the convergence rate can be dominated by $\frac{1}{\sqrt{PK}}$ and achieve linear speedup $O(\frac{1}{\sqrt{PK}})$.

\section{Experiment}
\subsection{Experimental Setting}

\textbf{Datasets and models:} The performance of all algorithms is evaluated in multiple deep learning tasks including image classification on CIFAR-10 and CIFAR-100~\citep{krizhevsky2009learning}. All training datasets are evenly partitioned over a network of workers (each client has all classes and the number of samples per classes are the same across all the clients).

\textbf{Compared algorithms:} We implement the proposed Non-blocking SGD on the state-of-the-art algorithms D-PSGD and MATCHA with a communication budget $c_b = 0.5$. MATCHA allows the system designer to set a flexible communication budget $c_b$, which represents the average frequency of communication over the links in the network. When $c_b$ = 1, MATCHA reduces to vanilla decentralized SGD. When we set $c_b < 1$ , MATCHA carefully reduces the communication frequency of each link, depending upon its importance in maintaining the overall connectivity of the graph. In addition, MATCHA assign probability to connections between workers so they may become active in some iterations. By using this disjoint links, MATCHA reduces the communication but maintains the degree. The Non-blocking D-PSGD (Algorithm~\ref{Alg:2}) and Non-blocking MATCHA are compared with their baseline. 

\textbf{Machines/Clusters:} All the implementations are compiled with PyTorch and OpenMPI within mpi4py. We conduct experiments on a HPC cluster with 100Gbit/s infini-band network. In all of our experiments, we use rtx8000 GPU as a node. To simulate the heterogeneous environment, we randomly slow down some of the nodes. The slowdown nodes and the rate of the slowdown are different in each iteration to simulate the real-world situation. The slowest node takes twice as long as the fastest node to finish a single batch. The slowdown is implemented using the sleep function.

\textbf{Implementations:} All algorithms are trained for a sufficiently long time until convergence or onset of over-fitting. The learning rate is fine-tuned for the D-PSGD baseline and then used for all other algorithms. Learning rate decay is adjusted according to the number of finished mini-batches in the experiment. We set the initial learning rate as 0.8 and it decays by 10 after 100 and
150 epochs. The batch size per worker node is 64. The batch size of baseline and the mini-batch size of Non-blocking algorithm are the same.

\subsection{Result Analysis}
The base topology of D-PSGD and MATCHA is shown in Figure~\ref{Fig:1}(a). In Figure~\ref{Fig:3} we present the comparison of performance on the CIFAR-10 dataset and in Figure~\ref{Fig:4} we present the performance on CIFAR-100 dataset.  

\textbf{End-to-end performance:} We first validate that, under certain network configurations, Non-blocking D-PSGD and Non-blocking MATCHA converge in similar numer of epochs, to a solution that is of similar quality as centralized SGD in a heterogeneous environment. Table~\ref{Tab:1}, Figure~\ref{Fig:3} (a, d), and \ref{Fig:4} (a, d) illustrate that the final test accuracy of Non-blocking algorithms is similar (or even better) than the two baselines, D-PSGD and MATCHA. 

\begin{table}[h]
\centering
\begin{tabular}{lllllllll}
\multicolumn{1}{c|}{\textbf{Dataset}} &\multicolumn{1}{c|}{\textbf{Algorithms}} & \multicolumn{2}{|c|}{Baseline} &\multicolumn{3}{c}{\textbf{Nonblocking}}\\ \hline
\multicolumn{1}{c|}{CIFAR-10}              &\multicolumn{1}{c|}{D-PSGD}              & \multicolumn{2}{|c|}{0.925}     &\multicolumn{3}{c}{0.928}   \\ 
\multicolumn{1}{c|}{CIFAR-10}              &\multicolumn{1}{c|}{MATCHA}              & \multicolumn{2}{|c|}{0.931}     &\multicolumn{3}{c}{0.935}   \\
\multicolumn{1}{c|}{CIFAR-100}              &\multicolumn{1}{c|}{D-PSGD}              & \multicolumn{2}{|c|}{0.783}     &\multicolumn{3}{c}{0.792}   \\ 
\multicolumn{1}{c|}{CIFAR-100}              &\multicolumn{1}{c|}{MATCHA}              & \multicolumn{2}{|c|}{0.794}     &\multicolumn{3}{c}{0.791}   \\ \hline
\end{tabular}       
\caption{\label{tab:widgets} Averaged test accuracy on CIFAR-10 and CIFAR-100}
\label{Tab:1}
\end{table}

\textbf{Speedup: } As we discussed in theoretical analysis, the bottleneck of wall-clock epoch time in Non-blocking algorithms is the fastest worker instead of the slowest straggler in the baseline algorithms. The convergence run-time speed up is linear w.r.t. $\frac{E[X_{P:P}]}{E[X_{1:P}]}$. In a heterogeneous environment where the fastest learner runs two times faster than the slowest straggler, the Non-blocking SGD takes up to 2x fewer times to reach the same final training loss.

\subsection{Conclusion}
This paper proposes a Non-blocking stochastic gradient descent algorithm (Non-blocking SGD) to deal with the stragglers' problem. The algorithm is robust in heterogeneous environments by allowing GPUs to process different amounts of data in a single batch and minimizing the delays in waiting for the slowest learner. The Non-blocking algorithm is extendable to most of the-state-of-the-art baselines and any other distributed computation algorithm that requires frequent synchronizations. It is also theoretically justified to have the same convergence rate as its baseline and can achieve linear speedup w.r.t. $\frac{E[X_{P:P}]}{E[X_{1:P}]}$. Experiments using different baselines validate the proposed algorithm.

\newpage
\nocite{langley00}
\bibliography{example_paper}
\bibliographystyle{mlsys2023}

\clearpage
\newpage
\pagebreak
\appendix
\section{Expected run-time for Non-blocking Synchronous SGD under exponential distribution assumption}

\textbf{Remark 1} The mean of the exponential random variables are $\frac{1}{\lambda}$. The CDF of an exponential random variable with mean $\frac{1}{\lambda}$ is $1 - e^{-x\lambda}, x>0$. 
\textbf{Remark 2} Assume the wall clock time of each learner to process a single mini-batch be i.i.d. exponential random variables $X_1,X_2, . ,X_i, . . ,X_P \sim exp(\lambda)$. Each learner run-time distributed with common density $\lambda e^{-\lambda x}$, then $E(T_{Nonblocking}) = \min \left\{ X_1, X_2, ..., X_P\right\} \sim \lambda Pe^{-\lambda Px}$.
\newline\textbf{Remark 3} $P\left\{Y > y\right\} = 1 - P\left\{Y \leq y\right\}$. 
\newline\textbf{Proof of Theorem 2.1} According to Remark 2,3,4. We assume $T_{Nonblocking} = Y$.

$$
\begin{aligned}
E[T_{Nonblocking}] &= F_Y(Y) = P\left\{Y\leq y \right\} \\ 
&= P\left\{ \min \left\{  X_1,..., X_P\right\} \leq y \right\} \\
&= 1 - P\left\{ \min \left\{  X_1,..., X_P\right\} > y \right\} \\
&= 1 - P\left\{ X_1 > y \right\}...*P\left\{ X_P > y \right\} \\
&= 1 - e^{-y*\lambda}*e^{-y*\lambda}*...*e^{-y*\lambda} \\
&= 1 - e^{-y*\lambda}*e^{-y*\lambda}*...*e^{-y*\lambda} \\
&= 1 - e^{-P*y*\lambda}
\end{aligned}
$$

Then the density function will be the same as exponential distribution function,

$$
\begin{aligned}
f_Y(y) &= \frac{d}{dy}F_Y(Y) \\
f_Y(y) &= \lambda*P e^{-P*\lambda*y} \\
E[Y] &= \frac{1}{\lambda*P}
\end{aligned}
$$

\section{Expected run-time for vanilla Centralized SGD under exponential distribution assumption}
\textbf{Proof of Theorem 2.2} The wall clock time of each learner to process a single mini-batch be i.i.d. exponential random variables $X_1,X_2, . ,X_i, . . ,X_P \sim exp(\lambda)$. The expected maximum number of P learners under exponential distribution is $E(T_{Sync}) = \max \left\{ X_1, X_2, ..., X_P\right\}$.

$$
\begin{aligned}
&P(\max \left\{ X_1, X_2, ..., X_P\right\} \leq y) \\
&= P(X_1 \leq y)P(X_2 \leq y)*...*P(X_P \leq y) \\
&= (1-e^{-\lambda y})^P \\
\end{aligned}
$$
Hence, the PDF function of $y$ is,
$$
\begin{aligned}
&\frac{dP(\max \left\{ X_1, X_2, ..., X_P\right\} \leq y)}{dy} = \frac{d(1 - e^{-\lambda y})^P}{dy} \\
&= P*\lambda*e^{-\lambda y}*(1-e^{-\lambda y})^{P-1}
\end{aligned}
$$
Then we can get the expectation of $E(T_{Sync})$ is,
$$
\begin{aligned}
\int_{0}^{+\infty} \lambda Pye^{-\lambda y}(1-e^{-\lambda y})^{P-1} dy
\end{aligned}
$$
let $u = \lambda y, du = \lambda dy, y = \frac{u}{\lambda}$
$$
\begin{aligned}
& \int_{0}^{+\infty} \lambda Pye^{-\lambda y}(1-e^{-\lambda y})^{P-1} dy \\
&=\int_{0}^{+\infty}\lambda P\frac{u}{\lambda}e^{-u}(1-e^{-u})^{P-1}\frac{1}{\lambda}du \\
&=\int_{0}^{+\infty}Pue^{-u}(1-e^{-u})^{P-1}\frac{1}{\lambda}du \\
&=\frac{P}{\lambda}\int_0^{+\infty}ue^{-u}(1-e^{-u})^{P-1}du \\
\end{aligned}
$$
According to Bernoulli's theorem, $(1-e^{-u})^{P-1} = \sum_{k=0}^{P-1}\binom{P-1}{k}e^{-ku}(-1)^k$,
$$
\begin{aligned}
&\frac{P}{\lambda}\int_0^{+\infty}ue^{-u}(1-e^{-u})^{P-1}du \\
&= \frac{P}{\lambda}\int_0^{+\infty}ue^{-u}\sum_{k=0}^{P-1}\binom{P-1}{k}e^{-ku}(-1)^k du \\
&=\frac{P}{\lambda}\sum_{k=0}^{P-1}\binom{P-1}{k} (-1)^k\{ \left[-u\frac{1}{k+1}e^{-(k+1)u}\right]_0^{+\infty}\\
&\quad+\int_0^{+\infty}\frac{e^{-(k+1)u}}{k+1}du\} \\
&=\frac{P}{\lambda}\sum_{k=0}^{P-1}\binom{P-1}{k}(-1)^k\int_0^{+\infty}e^{-(k+1)u}du \\
&=\frac{P}{\lambda}\sum_{k=0}^{P-1}\binom{P-1}{k}\frac{(-1)^k}{k+1}\int_0^{+\infty}e^{-(k+1)u}du \\
&=\frac{P}{\lambda}\sum_{k=0}^{P-1}\binom{P-1}{k}\frac{(-1)^k}{(k+1)^2} \\
\end{aligned}
$$
The result are verified by program based on $P=10, N=10000$, the final result is the same as $\frac{1}{N}\sum_{i=0}^{N}max_i\left\{X_1, X_2, ...X_P\right\}$


\end{document}